%% file: main.tex
\documentclass[letterpaper, 10 pt, conference]{IEEEtran}  

\IEEEoverridecommandlockouts                              

\usepackage{amsmath} 
\usepackage{amssymb}  
\usepackage{bm}
\usepackage{multirow}
\usepackage{makecell}
\usepackage{bbding}

\usepackage{graphics} 
\usepackage{epsfig} 
\let\labelindent\relax

\usepackage{mathtools}
\usepackage{enumitem}
\usepackage{times} 

\usepackage[utf8]{inputenc} 
\usepackage[T1]{fontenc}    
\usepackage{hyperref}       
            
\usepackage{url}            
\usepackage{booktabs}       
\usepackage{amsfonts}       
\usepackage{nicefrac}       
\usepackage{microtype}      
\usepackage{xcolor}         
\usepackage{cleveref}
\usepackage{hyperref}
\usepackage{graphicx}
\usepackage{float}
\usepackage{multirow}
\usepackage{diagbox}

\usepackage{algorithm}
\usepackage{algpseudocode}
\usepackage{arydshln}
\usepackage{marvosym}

\usepackage{multirow}
\usepackage{listings}
\usepackage{graphicx}
\usepackage{siunitx} 
            
\definecolor{dkgreen}{rgb}{0,0.6,0}
\definecolor{gray}{rgb}{0.5,0.5,0.5}
\definecolor{mauve}{rgb}{0.58,0,0.82}
\title{\LARGE \bf
GaussianGrasper: 3D Language Gaussian Splatting for Open-vocabulary Robotic Grasping
}

\author{Yuhang Zheng, Xiangyu Chen, Yupeng Zheng, Songen Gu, Runyi Yang, Bu Jin, Pengfei Li,
Chengliang Zhong, \\ Zengmao Wang, Lina Liu, Chao Yang, Dawei Wang, Zhen Chen, Xiaoxiao Long$^{\ast}$, Meiqing Wang$^{\ast}$
\thanks{Yuhang Zheng is with the School of Mechanical Engineering and Automation, Beihang University and EncoSmart.
Xiangyu Chen and Zhen Chen are with the EncoSmart.
Yupeng Zheng and Bu Jin are with the Institute of Automation, Chinese Academy of Sciences.
Songen Gu, Pengfei Li and Chengliang Zhong are with the AIR, Tsinghua University.
Runyi Yang is with the Imperial College London.
Lina Liu is with the China Mobile Research Institute.
Zengmao Wang is with the School of Computer Science, Wuhan University.
Chao Yang is with the Shanghai AI Laboratory.
Dawei Wang and Xiaoxiao Long are with the Department of Computer Science, the University of Hong Kong.
Meiqing Wang is with the School of Mechanical Engineering and Automation, Beihang University.}
\thanks{\textbf{$^{\ast}$ Xiaoxiao Long and Meiqing Wang are the corresponding authors. \protect\\ Email: xxlong@connect.hku.hk, wangmq@buaa.edu.cn}}
}

\begin{document}
\maketitle


\input{a_abstract}


\input{b_introduction}
\input{c_related_work}
\input{d_method}
\input{e_experiment}
\input{f_limitation}
\input{g_conclusion}

\bibliographystyle{IEEEtran}
\bibliography{IEEEabrv,reference}

\end{document}

%% file: a_abstract.tex
\begin{abstract}

Constructing a 3D scene capable of accommodating open-ended language queries, is a pivotal pursuit, particularly within the domain of robotics. Such technology facilitates robots in executing object manipulations based on human language directives. To tackle this challenge, some research efforts have been dedicated to the development of language-embedded implicit fields. However, implicit fields (e.g. NeRF) encounter limitations due to the necessity of processing a large number of input views for reconstruction, coupled with their inherent inefficiencies in inference.
Thus, we present the \textit{GaussianGrasper}, which utilizes 3D Gaussian Splatting to explicitly represent the scene as a collection of Gaussian primitives. Our approach takes a limited set of RGB-D views and employs a tile-based splatting technique to create a feature field. In particular, we propose an Efficient Feature Distillation (EFD) module that employs contrastive learning to efficiently and accurately distill language embeddings derived from foundational models. With the reconstructed geometry of the Gaussian field, our method enables the pre-trained grasping model to generate collision-free grasp pose candidates.
Furthermore, we propose a normal-guided grasp module to select the best grasp pose.
Through comprehensive real-world experiments, we demonstrate that \textit{GaussianGrasper} enables robots to accurately query and grasp objects with language instructions, providing a new solution for language-guided manipulation tasks. Data and codes can be available at \href{https://github.com/MrSecant/GaussianGrasper}{https://github.com/MrSecant/GaussianGrasper}.

\end{abstract}


%% file: b_introduction.tex
\section{Introduction}
Recently, there has been an increasing scholarly focus on language-guided robotic manipulation due to its vast potential in facilitating human-robot interaction, enabling robotic home services, and enhancing flexible manufacturing.
Imagine that a robot is asked to pick up the water cup in a cluttered, unstructured environment, it needs to (1) locate the water cup via responding to language description; (2) be aware of the geometry to execute a stable grasp. 
In this process, understanding the diverse objects with different shapes and material properties in the open world is the pivotal challenge.

\begin{figure}
  \centering
  \includegraphics[width=\linewidth]{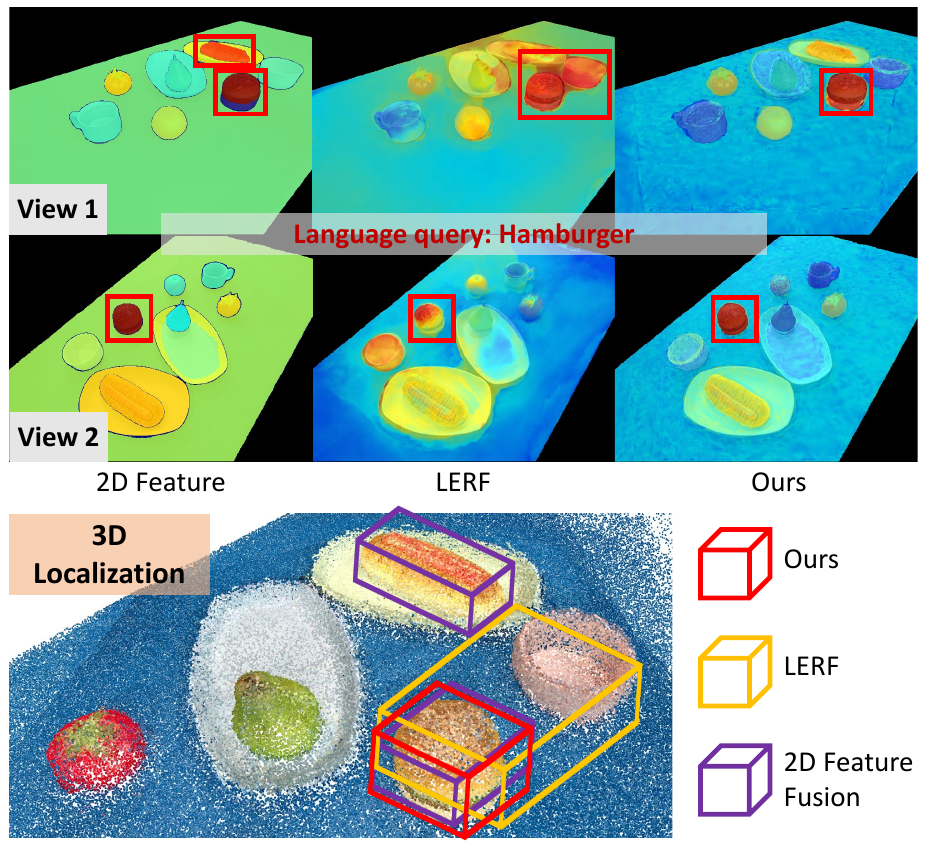}
  \vspace{-5mm}
  \caption{We present a comparison between our method, 2D feature fusion, and LERF. When given the language query "hamburger", the features extracted by the 2D foundation models exhibit inconsistencies between two viewpoints, and LERF lacks clear segmentation boundaries. Consequently, they both suffer from imprecise 3D localization, as depicted by the yellow and purple 3D bounding boxes. In contrast, our method reconstructs a consistent feature field and achieves more precise 3D localization.}
  \label{fig:teaser}
  \vspace{-4mm}
\end{figure}

Although many works have proposed various solutions, existing capabilities of scene understanding are insufficient to afford language-guided manipulation. Most existing works are based on 2D images~\cite{Lynch2020LanguageCI, shridhar2022cliport, lin2023spawnnet, guhur2023instruction} which are efficient but have limitations for robotic manipulation as robots can not easily infer visual occlusion and spatial relation from multi-view misaligned images. 

To obtain precise 3D positions for robotic manipulation, recent works have focused on 3D representations. A straightforward approach is leveraging 2D visual models~\cite{kirillov2023segment, radford2021learning, caron2021emerging} to extract semantics and then fuse the 2D semantics into 3D points or volume. However, this fusion strategy suffers from semantic inconsistency in 3D, as the semantics provided by the visual model are not consistent across multi-views.
Other methods~\cite{chen2020scanrefer, jiang2021synergies, Chen2023PolarNet3P, shridhar2023perceiver, ze2023gnfactor, Zhong_2023_ICCV} that use 3D backbone to extract features and are supervised by 3D annotation or manipulation feedback can effectively make robots explicitly understand 3D scenes and learn skills but are difficult to apply to the real world due to the challenges in data acquisition and annotation.
Recently, distilled feature fields (DFFs)~\cite{kobayashi2022decomposing, tschernezki2022neural, kerr2023lerf} which reconstruct 3D feature fields from 2D images via implicit representation were introduced. 
Based on 2D-to-3D distillation, recent works~\cite{shen2023distilled, rashid2023language} have made impressive progress in improving 3D scene understanding and enabling robots to interact with the physical world according to natural language. 
However, DFFs can not be inflexibly applied for robotics manipulation as most of these methods suffer from (1) imprecise localization as these methods extract patch-level features, resulting in ambiguous boundaries; (2) high costs of collecting dense training views (e.g. 50 views in F3RM~\cite{shen2023distilled}); (3) slow inference speed, hindering robots from responding to language instructions in time and (4) weak ability to cope with scene changes caused by manipulation.

To tackle problems, we introduce \textbf{GaussianGrasper}, an open-world robotic manipulation system based on 3D Gaussian Splatting (3DGS)~\cite{kerbl20233d}, which models the 3D scene as a set of 3D Gaussian primitives. 
Our insight is that we (1) reconstruct a 3D feature field via efficient feature distillation to support language-guided localization; (2) render depth and surface normal to provide detailed local geometry, enabling the generation of feasible grasping poses; (3) operate Gaussian primitives and fine-tune 3DGS to update the changed scene.

More specifically, our method enables language-guided manipulation via the following steps: 
(1) Initialization: we scan RGB-D images of a few viewpoints to initialize the 3DGS, reducing the cost of data collection. 
(2) Feature field reconstruction: we propose an efficient feature distillation (EFD) module that employs SAM~\cite{kirillov2023segment} and CLIP~\cite{radford2021learning} to extract dense and shape-aware 2D descriptors and leverage contrastive learning to efficiently optimize the distilled features. 
(3) Localization and grasp: we use open-vocabulary queries to locate the target object and use a pre-trained grasping module to provide grasp poses where rendered normal is used to filter out unfeasible proposals based on \textit{Force-closure} theory.  
(4) Scene updating: After executing manipulation, we update the scene by operating corresponding Gaussian primitives and fine-tuning 3DGS with images from fewer views.

In summary, the contributions of this paper are as follows:

\begin{itemize}
\item[$\bullet$] We introduce \textbf{GaussianGrasper}, a robot manipulation system implemented by a 3D Gaussian field endowed with open-vocabulary semantics and accurate geometry that is capable of rapid updates to support open-world manipulation tasks guided by language.

\item[$\bullet$] We propose EFD that leverages contrastive learning to efficiently distill CLIP features and augment feature fields with SAM segmentation prior, addressing computational expense and boundary ambiguity challenges.

\item[$\bullet$] We propose a normal-guided grasp module that uses rendered normal to filter out unfeasible grasp poses. 

\item[$\bullet$] We demonstrate the system's zero-shot generalization capability for manipulation tasks in multiple real-world household tabletop scenes and common objects.
\end{itemize}

%% file: c_related_work.tex
\section{Related Work}

\subsection{Grasp Pose Detection}
Grasp pose detection is the pivotal part of robot grasping, which plays a critical role in enabling the robot to interact with objects in the physical world. Previous 3-DoF Grasping methods treated the grasping task as 2D pose detection~\cite{redmon2015real, Mahler2017DexNet2D, guo2017hybrid, zhou2018fully, zheng2023enhancing}. They define the grasp pose as a  fixed-height oriented rectangle and predict the orientation and the width of the rectangle. However, their predicted grasp poses are limited to 3-DoF due to the lack of 3D geometry. To allow robots to plan higher dexterous grasps, extensive works focuses on 6-DoF grasping which use depth to augment the grasp pose detection~\cite{zhu20216, zhai2023monograspnet} or leverage point cloud as input to provide local geometry~\cite{liang2019pointnetgpd, wu2020grasp, sundermeyer2021contact, zhao2021regnet, alliegro2022end}. These methods exhibit high success rate when the depth is accurate but suffering from performance drops when encountering photometrically challenging objects such as transparent objects. To further improve the performance, some work fuses RGB with depth~\cite{fang2020graspnet, fang2022transcg, fang2023anygrasp} as RGB can alleviate the depth-missing problem.

In this paper, we use the RGB-D based method to generate grasp poses. To achieve stable grasp, we further explicitly utilize \textit{Force-closure} theory~\cite{ten2018using} to enhance grasp pose detection where estimated normal is used to filter out unfeasible grasp poses (more details in \textit{Normal-guided grasp} of Sec.~\ref{subsec: normal_filter}). 


\subsection{Reconstructing 3D Feature Field For Manipulation}
A number of recent work integrate 2D foundation models with 3D feature fields in contexts other than robotic manipulation~\cite{tung2019learning, kobayashi2022decomposing, tung20203d, peng2023openscene, kerr2023lerf, wang2022clip}. Based on the implicit representation, methods~\cite{kobayashi2022decomposing, wang2022clip, kerr2023lerf} leverage feature distillation via neural rendering to reconstruct 3D feature field. While as explicit representations, methods~\cite{tung2019learning, tung20203d, peng2023openscene} re-project 2D features and optimize the feature field in 3D.

For robot manipulation, recent works like F3RM~\cite{shen2023distilled} and LERF-TOGO~\cite{rashid2023language} leverage feature distillation to improve the robot understanding of 3D scenes and enable language-guided grasping. 
However, these methods need images from many viewpoints as input and can not quickly update the scene to cope with changes (movement or rotation) of objects.
As an explicit representation, SparseDFF~\cite{wang2023sparsedff} re-projects 2D features to 3D and leverages fusion strategy to optimize the feature field, leading to a significant reduction in usage of the number of viewpoints. 
However, explicit representation is not efficient for carrying high-dimension language features due to memory usage and computation time.

Different from the methods above, we propose an efficient feature distillation module based on explicit 3D Gaussians representation which uses segmentation priors provided by SAM to speed up feature field reconstruction and reduce the usage of memory (more details in Sec.~\ref{subsec: DFF}).

%% file: d_method.tex
\begin{figure*}
  \centering
  \includegraphics[width=0.99\textwidth]{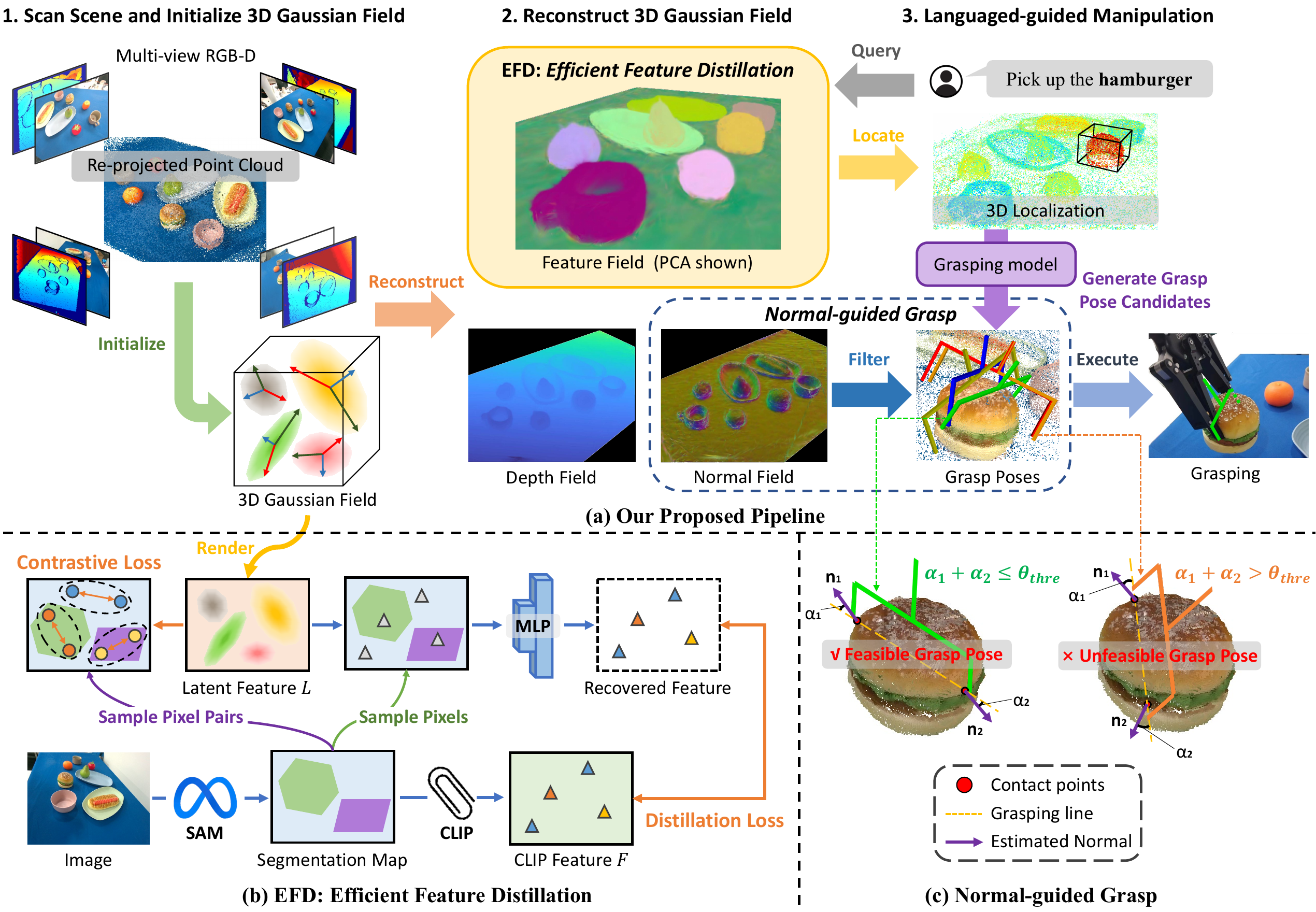}
  \vspace{-2mm}
  \caption{The architecture of our proposed method. (a) is our proposed pipeline where we scan multi-view RGBD images for initialization and reconstruct 3D Gaussian field via feature distillation and geometry reconstruction. Subsequently, given a language instruction, we locate the target object via open-vocabulary querying. Grasp pose candidates for grasping the target object are then generated by a pre-trained grasping model. Finally, a normal-guided module that uses surface normal to filter out unfeasible candidates is proposed to select the best grasp pose. (b) elaborates on \textit{EFD} where we leverage contrastive learning to constrain rendered latent feature $L$ and only sample a few pixels to recover features to the CLIP space via an MLP. Then, the recovered features are used to calculate distillation loss with the CLIP features. (c) shows the normal-guided grasp that utilizes Force-closure theory to filter out unfeasible grasp poses.}
  \label{fig:pipeline}
\end{figure*}

\section{Methodology}
In general terms, our method aims to pick up objects or place objects in specified locations according to the language instructions. 
The pipeline is shown in Fig.~\ref{fig:pipeline} (a) where our method (1) collects multi-view RGB-D images as input to initialize 3D Gaussian field; (2) reconstructs 3D feature field via efficient feature distillation module and (3) achieves languaged-guided manipulation. Specifically, we first introduce how to initialize 3DGS and the differentiable rasterizer of 3DGS in section~\ref{subsec: 3DGS}. Next, we elaborate on the \textit{EFD} module in section~\ref{subsec: DFF}. Finally, we introduce how to achieve language-guided manipulation in detail in section~\ref{subsec: grasp}, including locating objects through language queries, using a pre-trained grasping model to generate grasp poses, proposing a normal-guided grasp strategy to select feasible poses and updating the scene after manipulation.

\subsection{Preliminaries: 3D Gaussian Splatting}
\label{subsec: 3DGS}
\noindent
\subsubsection{Gaussian Primitive Initialization}
\label{subsec: init}
3DGS uses Structure from Motion (SfM)~\cite{schonberger2016structure} as the initialization part, which takes a collection of RGB images as input and outputs a sparse point cloud. 
These points construct a set of Gaussian primitives, each defined by mean \textbf{$\mu$} and a 3D covariance matrix $\Sigma= RSS^TR^T$, where $R$ and $S$ represent the rotation matrix and the scaling matrix.
To reduce the viewpoints and speed up initialization, we employ multi-view depth to re-project pixels of RGB images into the world coordinate to initialize Gaussian field.
Each point is the center of a Gaussian primitive and the rotation matrix and scaling matrix are initialized randomly.

\subsubsection{Differentiable Rasterizer for 3D Gaussians}
\label{subsec: Rasterizer}
3DGS renders the Gaussian primitives into images in a differentiable manner to optimize the parameters. 
Given a set of 3D Gaussian primitives $\mathcal{G}=\{g_i \mid i=1,2,3,..,n\}$, each 3D Gaussian primitive $g_i$ is first projected onto the corresponding 2D plane and is then sorted based on its depth $d_i$ from the viewpoint plane.
By tile-based rasterization, we sum up the pixel color $C(u)$ after sorting each primitive:
\begin{equation}
    C(u) = \sum_{i} c_i\alpha_i \prod_{j=1}^{i-1} (1-\alpha_j)
\label{render}
\end{equation}
where $c_i$ is a feature vector represented by spherical harmonics (SH) and $\alpha_i$ is obtained by multiplying Gaussian weight with opacity $\alpha$ associated to Gaussian primitives.

\subsection{Efficient Feature Distillation}
\label{subsec: DFF}
We reconstruct 3D open-vocabulary feature field via extracting dense CLIP features and efficiently distilling them into 3D, based on 3DGS.
The open-vocabulary reconstruction enables the scene to respond to natural language instructions.    

\subsubsection{Instance-level Segmentation Prior and Open-vocabulary Features}
To exact dense and shape-aware open-vocabulary features, we first use SAM to produce a set of instance-level masks. Then, we leverage CLIP to obtain open-vocabulary features for each mask.
Concretely, we process the input images through SAM to obtain a set of mask proposals and corresponding scores. Based on these scores, a non-maximum suppression strategy~\cite{neubeck2006efficient} is then implemented to filter superfluous masks. The resultant filtered set of masks constitutes a segmentation map of the image, representing instance-level priors. After filtration, we process each valid mask-aligned image region into the CLIP model to extract open-vocabulary features. Finally, we incorporate CLIP features of all masks into a feature map, referred to as $F$. 

\vspace{1mm}
\subsubsection{Open-vocabulary Feature Distillation}
We propose a novel and efficient open-vocabulary feature distillation method that starts with enhancing each 3D Gaussian primitive with embedded open-vocabulary feature.
As an explicit representation, a 3D Gaussian field can be composed of millions of primitives. Directly incorporating high-dimensional CLIP features (over 500 dimensions) into all primitives will result in unacceptable memory costs and computation time. 
Therefore, we compress the embedded open-vocabulary feature of Gaussian primitives from high-dimension CLIP space to low-dimension latent space. 
After initializing 3D Gaussian primitives with low-dimension latent feature $l$, we employ a feature rasterizer to render the feature map $L$:
\begin{equation}
    L(u) = \sum_{i} l_{i}\alpha_i \prod_{j=1}^{i-1} (1-\alpha_j)
\label{fea_render}
\end{equation}
where $l_{i}$ is the open-vocabulary feature embedding of $i^{th}$ 3D gaussian primitives and $L(u) $ represents the rendered open-vocabulary feature embedding at pixel $u$.

To distill the 2D open-vocabulary feature to 3D field, we need to (1) recover the dimension of $L$ to that of $F$ and (2) minimize the feature distance between the recovered $L$ and the $F$. 
However, high-dimensional vector computation for dense feature maps leads to a catastrophic increase in computation time and memory usage. 
To tackle this problem, we propose a contrastive-learning-based distillation strategy, as shown in Fig.~\ref{fig:pipeline} (b).
Specifically, having instance-level segmentation masks extracted by SAM, we impose constraints for the consistency of each pixel's rendered feature within the same mask. 
To ensure runtime efficiency, we randomly sample a few pixel pairs within each mask and only minimize the distance of latent features between pixels of per sampled pair. 
The number of pairs for each mask is proportion to the mask's area and the number of total pairs $n$ is fixed.
We calculate the average feature distances between all pairs as contrastive loss, written as:
\begin{equation}
    \mathcal{L}_{contr.} = 1 - \frac{1}{n}\sum_{i=1}^{n} L(u_i) \cdot L(v_i)
\end{equation}
where $u_i$ and $v_i$ are pixels in the $i^{th}$ sampled pair.
As the contrastive loss homogenizes the features within each mask, we only need to recover latent features of per mask to the CLIP space and subsequently minimize the distance between the recovered feature and the CLIP feature. 
In practice, we randomly sample the same number of pixels within each mask, whose latent features are then recovered via a trainable decoder $\Psi$ composed of two fully connected layers. We calculate the distillation loss between the recovered feature and the CLIP feature of all sampled $k$ pixels, defined as:
\begin{equation}
    \mathcal{L}_{distill} = 1 - \frac{1}{k}\sum_{i=1}^{k}\Psi(L(i)) \cdot F(i)
\end{equation}

By enhancing the low-dimension open-vocabulary feature embedding of 3D Gaussian and using contrastive learning which leverages the segmentation prior derived from SAM, our method provides a powerful and efficient solution for reconstructing 3D open-vocabulary representation. 

\subsection{Language-guided Robotic Manipulation}
\label{subsec: grasp}
We use the reconstructed feature field to conduct robotic manipulation. Given a language instruction, our method begins with employing open-vocabulary queries to locate the target object. Subsequently, we render the depth and normals of 3D Gaussian primitives to obtain the object's detailed geometry. Then, a point cloud-based grasping module is used to generate grasp poses and the rendered normal is used to filter out unfeasible ones. After manipulating objects, we quickly update the scene using observations from fewer viewpoints.

\subsubsection{Open-vocabulary Querying}
\label{subsubsec: query}
As our reconstructed feature field is aligned with natural language, we can locate the object described by language instructions via open-vocabulary querying.
We first follow the approach of LERF~\cite{kerr2023lerf} to compute the relevance score $s$ for each textual query: 
\begin{equation}
    s = \min_i\frac{\exp(\Psi(L)\cdot T^{q})}{\exp(\Psi(L)\cdot T^{q}) + \exp(\Psi(L)\cdot T_{i}^{canon})}
\end{equation}
where $T^{q}$ is the CLIP embedding of the text query and $T_{i}^{canon}$ is a set of canonical phrase embeddings $T_{i}^{canon}$ selected from "object", "things", "stuff", and "texture".
 
As a result, for each textual query, we obtain a relevance heatmap where the points with relevance scores below a predetermined threshold will be filtered out. Thus, the remaining region forms a mask for predicting the queried object. After obtaining the mask of the queried object, we locate the object with a bounding box and convex hull.

\subsubsection{Geometry Restruction}
To obtain dense point cloud representations of objects and surface normal which is closely related to robotic grasping, we render depth and surface normal to multiple viewpoints.

\noindent
\textit{Depth rendering:} Similar to the rendering of RGB, we compute the depth value for each pixel using the rasterizer:
\begin{equation}
    D(u) = \sum_{i} d_i\alpha_i \prod_{j=1}^{i-1} (1-\alpha_j)
\end{equation}
where $D(u)$ indicates the rendered depth map at pixel $u$.

We use the depth map obtained from the depth camera for the corresponding viewpoint to supervise the rendered depth map where we calculate the L1 loss:
\begin{equation}
    \mathcal{L}_{depth} = \frac{1}{m}\sum_{i=1}^{m} \lvert \hat{D}(i) - D(i) \rvert
\end{equation}
where $\hat{D}$ is the depth map obtained from the depth camera and $m$ is the number of pixels with valid depth value.

\noindent
\textit{Normal rendering:} As surface normals are directional vectors that should exhibit rotational equivariance, normals cannot be rendered as semantic features.
Therefore, we follow~\cite{jiang2023gaussianshader, long2024adaptive} that use the shortest axis direction of the 3D Gaussian primitives to serve as surface normal. 
As a result, the normals are geometric properties of Gaussian primitives, related to their orientations. We render the normal map by using the rasterizer:
\begin{equation}
    \textbf{N}(u) = \sum_{i} \textbf{n}_i\alpha_i \prod_{j=1}^{i-1} (1-\alpha_j)
\end{equation}
where $\textbf{N}(u)$ indicates the rendered surface normal map at pixel $u$ and $\textbf{n}_i$ is the normal of the $i^{th}$ 3D Gaussian primitive.

The rendered normal map represents the per-pixel surface normal in the robot base coordinate. We use a Sobel-like operator to compute the normals for each pixel in the acquired depth map and transform the calculated normals from the camera coordinate to the robot base coordinate. After normalizing all normal maps to unit vectors, we supervise the rendered normals within the valid depth region:
\begin{equation}
    \mathcal{L}_{normal} = \frac{1}{m}\sum_{i=1}^{m} \left( \left( \hat{\textbf{N}}(i) - \textbf{N}(i) \right)^{2} + 1 - \hat{\textbf{N}}(i) \cdot \textbf{N}(i) \right)
\end{equation}

\subsubsection{Feasible Grasp Pose Generation}
After obtaining the localization and geometry of the object, we (1) employ a grasp detection method to propose initial grasp poses and (2) utilize our rendered normal to augment the grasp pose proposals according to the force closure theory.

\noindent
\textit{Grasp pose generation:} 
In our work, we employ the AnyGrasp~\cite{fang2023anygrasp}, the state-of-the-art grasp detection network, which takes colorful point cloud as input and generates a set of collision-free grasp proposals for parallel two-finger grippers. 
Each grasp proposal is represented by grasp position, width, height, depth and a grasping score.
We generate the grasp-pose proposals for the object queried by language with the following steps: 
(1) Re-project the rendered depth of the queried object from each viewpoint to produce a dense point cloud representation. 
(2) Combine the object's point cloud with the point cloud derived from 3D Gaussian primitives because the derived point cloud depicts the localization and approximate shape of other objects, which is beneficial for generating collision-free grasp poses.
(3) Employ AnyGrasp to generate grasp poses, which are restricted in the aforementioned bounding box, obtained from steps in \ref{subsubsec: query}.

\noindent
\textit{Normal-guided grasp:} 
\label{subsec: normal_filter}
Although AnyGrasp provides a "grasping score" for each grasp proposal, the grasp pose with the highest score may not be the suitable one for stable grasp. Thus, we explicitly utilize the force closure theory as a screening mechanism. For each grasp pose proposal with two contact points, we calculate the angle between the grasping line and the surface normal of each contact point and get the sum of two angles. If the sum of the two angles is less than or equal to a pre-defined threshold $\theta_{thre}$, we regard the grasp pose as a feasible proposal, otherwise it is unfeasible, as shown in Fig.~\ref{fig:pipeline} (c). We choose the pose with the highest "grasping score" in feasible proposals as the final grasp pose.

\subsubsection{Gaussian Field Updating}
As an explicit representation composed of 3D Gaussian primitives, it is easy to operate Gaussian primitives. Therefore, after manipulating an object, we will operate all Gaussian primitives within this object's convex hull with the same rotation and translation, which can be calculated from the transform of the end-effector of the robot arm. After operating 3D Gaussian primitives to a new position, we use fewer views and time to fine-tune 3D Gaussian field to update the scene. 

%% file: e_experiment.tex
\begin{figure*}
  \centering
  \includegraphics[width=\linewidth]{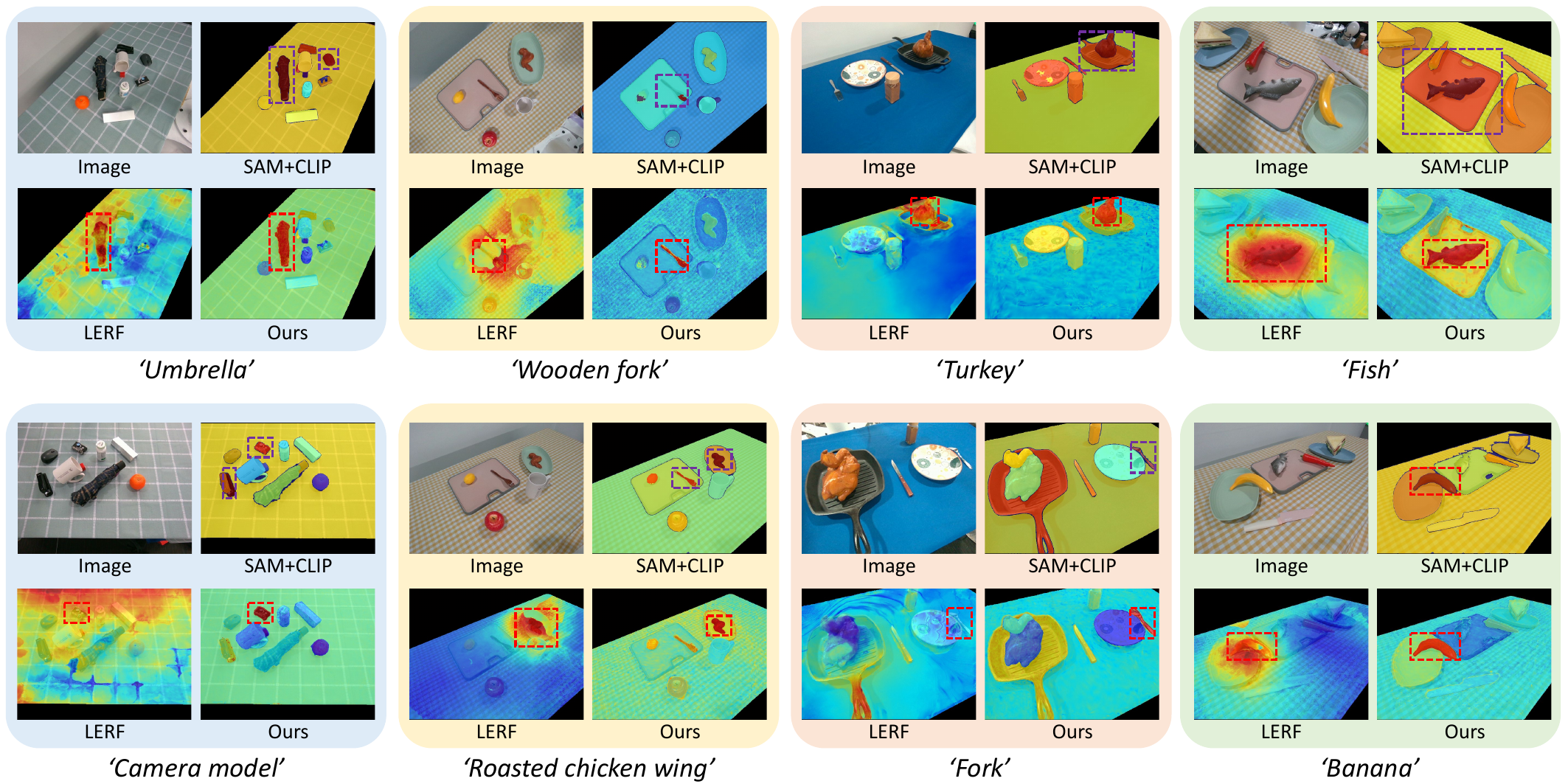}
  \vspace{-5mm}
  \caption{Relevance map of the given language instructions. Our method exhibits clearer segmentation boundaries compared to LERF, which can be used to obtain more accurate localization. Compared with SAM + CLIP, our approach exhibits more consistent open-vocabulary features across multi-views. For instance, in \textit{'Roasted chicken wing'}, the response of SAM + CLIP is the chicken wing and the fork while our method makes the correct response. }
  \label{fig:relevance}
  \vspace{-3mm}
\end{figure*}

\section{Experiment}
In this section, we first introduce the setup of the experimental environment. 
Next, we conduct experiments to validate our proposed EFD module where we report both quantitative results and qualitative results.
Subsequently, we show the results of geometry reconstruction and conduct ablation study to demonstrate the effectiveness of our proposed normal-guided grasp.
Finally, we conduct grasp-update-grasp experiments on scenes to verify the effectiveness and efficiency of our scene update module.
These experiments fully demonstrate the performance of our method in open-scene understanding and language-guided grasping.

\subsection{Experimental Setup}
\subsubsection{Scenes, Objects and Devices}
We built a $140\times70\times30\mathrm{cm}^{3}$ desktop scene with common objects in the kitchen including various food and tableware as well as office supplies including staplers, mouse and decorative ornaments. 
We use a UR5 robot arm equipped with a ROBOTIQ gripper to execute robotic manipulation.
We set up our system in 10 open desktop scenes with a total of 44 objects (40 are graspable) where we execute language-guided manipulation 120 times.
In terms of computing resources, we use an NVIDIA RTX-3090 GPU to reconstruct the feature field and reconstruct geometry. The reconstruction process only requires approximately 6GB of memory in total.

\subsubsection{Data Collection and Processing}
We first use the robot arm equipped with a Realsense D455 to scan the desktop scene from 16 viewpoints. At the same time, we will also record the camera extrinsic parameter, calculated through the transformation of the end effector to the base. After obtaining images and depth maps, we re-project the depth map and convert all re-projected points to the base coordinate. We downsample the number of these points to about 300k to initialize the 3D Gaussian primitives.
The collected images are processed by SAM and CLIP to generate segmentation maps and open-vocabulary feature maps. 

\subsection{Results of Efficient Feature Distillation}
We show both qualitative results and quantitative results to demonstrate the effectiveness and efficiency of our proposed EFD module. Our baselines are Lseg~\cite{li2022languagedriven} and LERF~\cite{kerr2023lerf} (All mention of LERF in our experiments includes an \textbf{extra depth supervision} to ensure a fair comparison with our method.)  

In qualitative results, we compare our method with SAM + CLIP (our 2D sudo labels) and LERF and show the relevance map of each given language instruction. As shown in Fig.~\ref{fig:relevance}, the purple boxes demonstrate that the features extracted from SAM and CLIP are not accurate such as the incomplete wooden fork in \textit{'Wooden fork'} and the pot and turkey with similar semantic features in \textit{'Turkey'}. In comparison, our relevance map is more accurate, proving that our reconstructed feature field solves the problem of feature inconsistency across multiple views. Besides, compared with LERF, our method exhibits better segmentation boundaries. Therefore, our method can help robots reduce the ambiguity of object perception.

We report the quantitative results of two tasks including segmentation and localization. 
In the segmentation task, as described in \ref{subsubsec: query} we filter out the region whose relevance score is below 0.85 to form a predicted segmentation map. We calculate the mIoU metric between predicted segmentation maps and our manually annotated ground truth.
In the localization task, following LERF, given a language instruction, if the point with the highest relevance score is in the target object, it is a successful localization. 
We calculate the average accuracy of all responses as the metrics. The results of segmentation and localization are shown in Table~\ref{tab:ablation2} where our method significantly outperforms other approaches. 

Besides, we test the query speed and report results in Table~\ref{tab:ablation2}. The metric is the time usage (s) per text query at a resolution of $640 \times 480$. It can be seen that our method achieves an approximate 180 × speedup over LERF. We also directly distill CLIP features into 3D Gaussian field, which takes over 70 GB of memory, making it hard to be applied to robots.

\begin{table}
    \centering
    \setlength{\tabcolsep}{0.03\linewidth}
    \caption{Quantitative comparisons of semantic segmentation and localization accuracy on our scenarios}
    \vspace{-2mm}
    \begin{tabular}{l | c c c}
        \hline
        {Method} & {mIoU(\%) $\uparrow$ } & {Accuracy(\%) $\uparrow$} & {Time per query (s) $\downarrow$}\\
        \hline
        LSeg\cite{li2022languagedriven} & 26.4 & 40.6 & {-} \\
        LERF*\cite{kerr2023lerf} & 41.3  & 65.1 & {40.27} \\
        Ours & \textbf{58.2} & \textbf{87.5} & {0.22} \\
        \hline
        \multicolumn{4}{l}{"*" represents LERF with an extra depth supervision} \\
    \end{tabular}
    \vspace{-3mm}
    \label{tab:ablation2}
\end{table}

\subsection{Results of Geometry Reconstruction}
\begin{figure}[h]
  \centering
  \includegraphics[width=\linewidth]{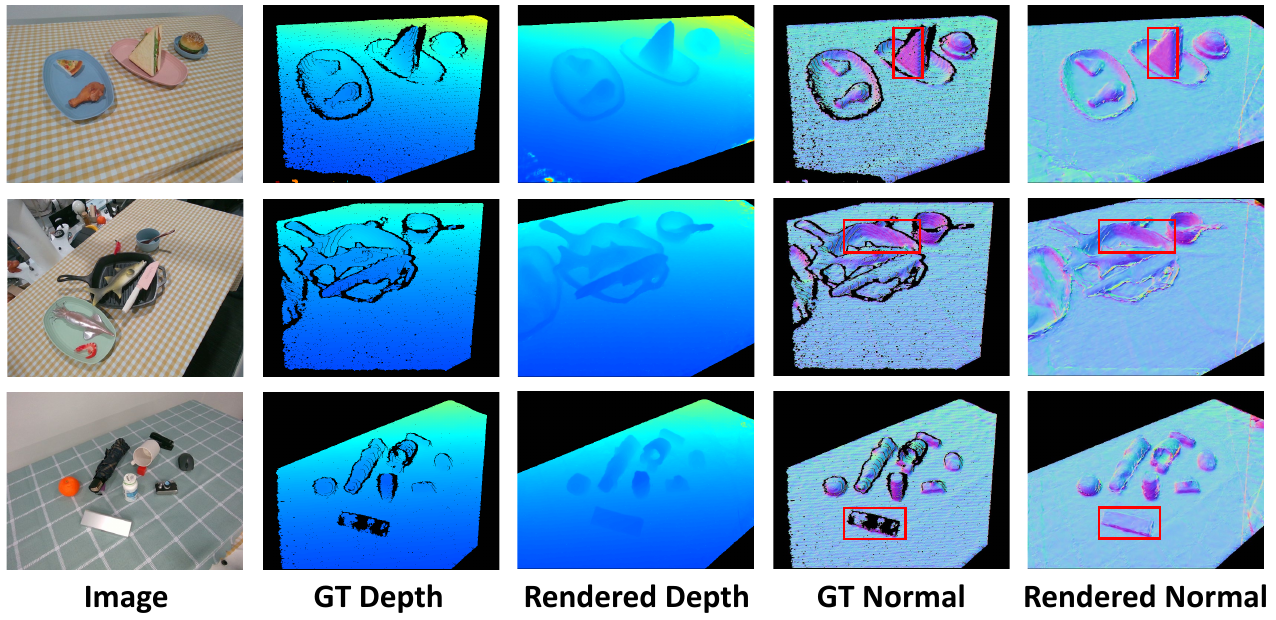}
  \vspace{-6mm}
  \caption{Compared with scanned depth and surface normal, our rendered depth and surface normal is smoother. Our method renders accurate depth and surface normal even in areas where the ground truth is invalid.}
  \label{fig:geometry}
\end{figure}

We show the visualization of our rendered depth and normal compared with ground truth (scanned by D455 camera), as shown in Fig.~\ref{fig:geometry}. The black region represents the invalid value of ground truth. It can be seen that the surface normal we rendered is smoother than the ground truth, as shown in red boxes. Furthermore, even in areas where the ground truth is invalid, we can still render accurate depth and surface normal. For example, in the third row, the camera can not capture the depth of the silver eyeglass case in this view because of the specular reflection of the case but our method renders accurate depth and normal of it. 

\subsection{Effectiveness of Normal-guided Grasp}
\begin{figure}[h]
  \centering
  \includegraphics[width=\linewidth]{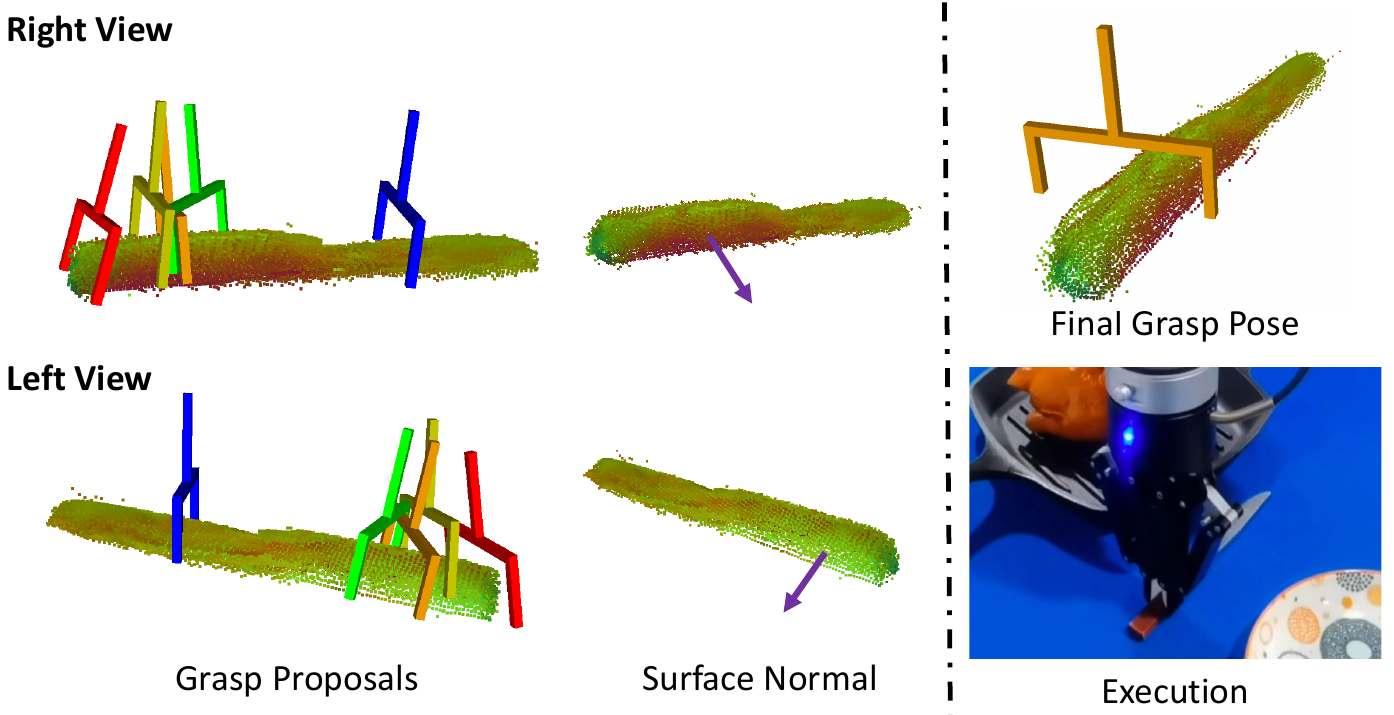}
  \vspace{-4mm}
  \caption{Effectiveness of our proposed normal-guided grasp. The left column shows the top 5 grasp proposals provided by AnyGrasp. The redder the color, the higher the grasping score. The middle column displays the surface normal of the object, with purple arrows indicating the normal of the contact points. The right column demonstrates the successful execution of grasping the knife utilizing the final grasp pose after filtering out unreasonable proposals.}
  \label{fig:normal}
  \vspace{-2mm}
\end{figure}

In this subsection, we aim to validate the effectiveness of our proposed normal-guided grasp. We first give the qualitative result to validate that the surface normal can filter out unfeasible grasp poses. As shown in Fig.~\ref{fig:normal}, the original top-ranked proposal (red) is filtered out as the angles between its grasping line and surface normal of contact points are too large. In contrast, the original second-ranked proposal is feasible. Thus, we execute a grasp based on this pose.

Besides, we report the quantitative results of the grasping success rate with and without the normal filter, as shown in Table~\ref{tab:grasp}. Leveraging the normal filter significantly increases the success rate by 7.7\%, further demonstrating the effectiveness of our proposed normal-guided grasp.

\begin{table}
    \centering
    \setlength{\tabcolsep}{0.03\linewidth}
    \caption{Grasping Results: Results are reported across 40 different objects of 10 scenarios. Each object is grasped three times.}
    \vspace{-2mm}
    \begin{tabular}{l|c}
        \hline 
        {Method} & {Grasping Success Rate (\%)} \\
        \hline
        LSeg + Depth\cite{li2022languagedriven} & 26.7 \\
        LERF + AnyGrasp\cite{kerr2023lerf} & 55.8 \\
        Ours w/o. Normal Filter & 78.3 \\
        Ours w/ Normal Filter & \textbf{85.0}  \\
        \hline
    \end{tabular}
     \vspace{-2.5mm}
    \label{tab:grasp}
\end{table}

\subsection{Results of Language-guided Manipulation}
In this subsection, we first give the result of language-guided grasping. Then, we validate our proposed scene updating via continuous language-guided picking and placing.   
\subsubsection{Successful rate of manipulation}
In this subsection, we show the result of languaged-guided grasping where we tested 120 times on 40 objects. We compare our method with (1) Lseg + AnyGrasp and (2) LERF + AnyGrasp. To obtain the 3D point cloud, we use the rendered depth to re-project the segmentation masks of LERF and use scanned depth to re-project the segmentation masks of LSeg. We define a successful grasp as stably picking up the corresponding object according to the language instruction and raising it to a height of more than 10cm over 3 seconds. The result is shown in Table~\ref{tab:grasp}, where our method far exceeds other methods in success rate.

\subsubsection{Scene updating}
\begin{figure}
  \centering
  \includegraphics[width=\linewidth]{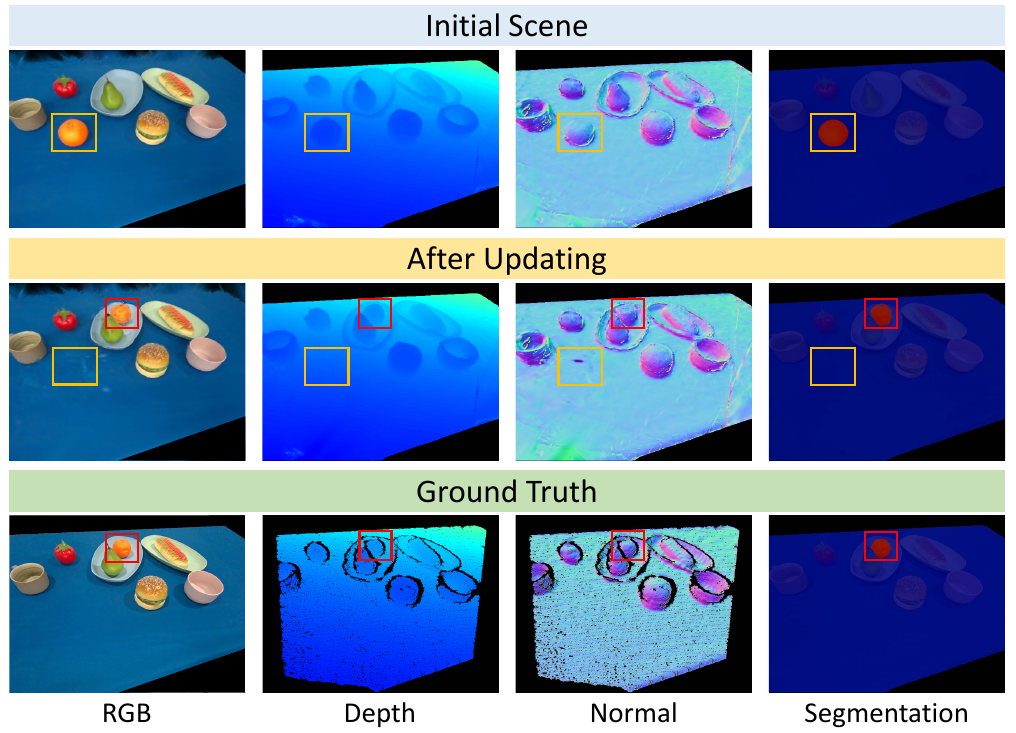}
  \vspace{-7mm} 
  \caption{Results of scene update. We show the RGB, depth, normal, and segmentation before and after the scene update based on the language query "orange". As indicated by the yellow boxes, our scene updating successfully moves the orange to the plate and restores the region that was previously obscured by the orange. As indicated by the red boxes, the updated orange still maintains accurate geometry and semantic features.}
  \label{fig:edit}
\end{figure}

To validate the effectiveness of our proposed efficient scene updating, we execute an experiment whose process is (1) picking up the object and placing it to the target position according to the language instruction, (2) capturing RGB-D images from 5 viewpoints to update the scene, (3) executing another manipulation on this object. The result is shown in Fig.~\ref{fig:edit}, where our updated scene remains high quality RGB, geometry and semantic features, proving the effectiveness of the proposed scene updating. Besides, we also show the efficiency comparison between LERF and ours including viewpoint numbers, memory usage and reconstruction time for updating, as shown in Table~\ref{tab:update}. Our scene update capability makes the reconstructed scene more capable of handling continuous grasping.

\begin{table}
    \begin{center}
    \setlength{\tabcolsep}{0.03\linewidth}
    \caption{Effieciency comparison between LERF and our method.}
    \vspace{-2mm}
    \begin{tabular}{l|c c c}
        \hline
        Method & 
        Viewpoints & Memory & Time \\
        \hline
        LERF~\cite{kerr2023lerf} & 16 & 15GB &30min \\
        Ours & \textbf{5} & \textbf{4GB} & \textbf{1min} \\
        \hline
    \end{tabular}
    \end{center}
     \vspace{-4mm}
    \label{tab:update}
\end{table}

%% file: f_limitation.tex
\section{Limitation} 

One limitation is that our reconstructed scene remains static. Although we have proposed a scene-updating module to handle continuous robotic grasping, we are unable to account for unrecordable scene changes, such as objects shifting positions caused by collisions or vibrations. 
Another limitation is that our method fails to estimate the depth and normal of transparent objects due to the lack of ground truth.

%% file: g_conclusion.tex
\section{Conclusion} 
\label{sec:conclusion}

In this paper, we introduce GaussianGrasper, a novel approach for open-world robotic grasping guided by natural language instructions from RGB-D inputs. Taking multi-view RGB-D images as input, our method efficiently reconstructs consistent feature fields through our proposed EFD module. The feature field enables the robot to understand the open world and make precise localization based on language instructions. Besides, we estimate the geometry and propose the normal-guided grasp to augment the robotic grasping.
Furthermore, our scene can also be quickly updated to support continuous grasping. 
Sufficient real-world experiments have demonstrated the effectiveness of our method.